\documentclass[letterpaper, 10 pt, conference]{ieeeconf}  % Comment this line out if you need a4paper

\IEEEoverridecommandlockouts                              % This command is only needed if
\let\labelindent\relax

\usepackage{graphics} % for pdf, bitmapped graphics files
\usepackage[pdftex]{graphicx} % for postscript graphics files
\usepackage{amsmath} % assumes amsmath package installed
\usepackage{amssymb}  % assumes amsmath package installed
\usepackage{blindtext}
\usepackage{todonotes}
\usepackage{acro}
\usepackage[normalem]{ulem}
\useunder{\uline}{\ul}{}
\usepackage{colortbl}
\usepackage{float}
\usepackage{balance}
\usepackage{cleveref}
\usepackage[inline]{enumitem}
\usepackage{siunitx}
\usepackage{mwe,tikz}
\usepackage[percent]{overpic}
\usepackage{subcaption}
\usepackage[font={footnotesize}]{caption}
\usepackage[inline]{enumitem}

\newlength{\subImageWidth}

\DeclareAcronym{2d}{short = 2D, long = two-dimensional}
\DeclareAcronym{ros}{short = ROS, long = Robot Operating System}
\DeclareAcronym{cnn}{short = CNN, long = convolutional neural network}
\DeclareAcronym{rnn}{short = RNN, long = recurrent neural network}
\DeclareAcronym{fc}{short = FC, long = fully-connected}
\DeclareAcronym{lstm}{short = LSTM, long = Long Short-Term Memory}
\DeclareAcronym{relu}{short = ReLU, long = Rectified Linear Units}
\DeclareAcronym{fov}{short = FOV, long = field of view}
\DeclareAcronym{dnn}{short = DNN, long = deep neural network}
\DeclareAcronym{uav}{short = UAV, long = unmanned aerial vehicle}
\DeclareAcronym{mpc}{short = MPC, long = model predictive controller}
\DeclareAcronym{dwa}{short = DWA, long = dynamic window approach}

\newcommand{\mat}[1]{\boldsymbol{\mathbf{#1}}}
\renewcommand{\u}{\mat{u}}
\newcommand{\y}{\mat{y}}
\newcommand{\g}{\mat{g}}
\newcommand{\params}{\mat\theta}
\newcommand{\F}{\mathcal{F}}
\newcommand{\training}{\textit{train} }
\newcommand{\evalsimple}{$\mathrm{\textit{eval}}_{1}$ }
\newcommand{\evalcomplex}{$\mathrm{\textit{eval}}_{2}$ }

\title{From Perception to Decision: A Data-driven Approach to End-to-end Motion Planning for Autonomous Ground Robots}
\author{Mark Pfeiffer, Michael Schaeuble, Juan Nieto, Roland Siegwart and Cesar Cadena%
\thanks{This work has partially received funding from
the European Union Seventh Framework Programme FP7/2007-2013, Challenge~2,
Cognitive Systems, Interaction, Robotics, under grant agreement No. 610603, EUROPA2.}
\thanks{The authors are with the
ETH Zurich, Zurich, Switzerland. \newline
{\tt \{pfmark, schamich, nietoj, rsiegwart, cesarc\}@ethz.ch}.}%
}

\begin{document}

\onecolumn
{\large
%\documentclass[letterpaper, 12 pt]{article}
%\setlength{\parindent}{0pt}
%\begin{document}

\thispagestyle{empty}

{\setlength{\parindent}{0cm}
\textcopyright 2017 IEEE. Personal use of this material is permitted. Permission from IEEE must be obtained
for all other uses, in any current or future media, including reprinting/republishing this material
for advertising or promotional purposes, creating new collective works, for resale or redistribution
to servers or lists, or reuse of any copyrighted component of this work in other works.
}
\\

{\setlength{\parindent}{0cm}
Pre-print of article that was presented at the 2017 IEEE International Conference on Robotics and Automation (ICRA).
}
\\

{\setlength{\parindent}{0cm}
Please cite this paper as:
}
\\

{\setlength{\parindent}{0cm}
M. Pfeiffer, M. Schaeuble, J. Nieto, R. Siegwart, C. Cadena. (2017). \\ "From Perception to Decision: A Data-driven Approach to End-to-end Motion Planning for Autonomous Ground Robots" in IEEE International Conference on Robotics and Automation (ICRA), 2017.
}
\\

{\setlength{\parindent}{0cm}
bibtex:
}
\\

{\setlength{\parindent}{0cm}
%@inproceedings\{gawel2018x-view,\\
%  \phantom{x} title=\{X-View: Graph-Based Semantic Multi-View Localization\},\\ 
%  \phantom{x} author=\{Gawel, Abel and Del Don, Carlo and Siegwart, Roland and Nieto, Juan and Cadena, Cesar\},\\ 
%  \phantom{x} booktitle={IEEE Robotics and Automation Letters (RA-L)\},\\
%  \phantom{x} year=\{2018\}
%  \\ 
%\}

@inproceedings\{pfeiffer2017perception, \\
  \phantom{x} title=\{From Perception to Decision: A Data-driven Approach to End-to-end Motion Planning for Autonomous Ground Robots\},\\
  \phantom{x} author=\{Pfeiffer, Mark and Schaeuble, Michael and Nieto, Juan and Siegwart, Roland and Cadena, Cesar\},\\
  \phantom{x} booktitle=\{IEEE International Conference on Robotics and Automation (ICRA)\},\\
  \phantom{x} pages=\{1527--1533\},\\
  \phantom{x} year=\{2017\},\\
  \phantom{x} organization=\{IEEE\}
  \\
\}

}

%\end{document}

\par}
\normalsize
\twocolumn
\clearpage
\setcounter{page}{1}

\maketitle
\thispagestyle{empty}
\pagestyle{empty}

%%%%%%%%%%%%%%%%%%%%%%%%%%%%%%%%%%%%%%%%%%%%%%%%%%%%%%%%%%%%%%%%%%%%%%%%%%%%%%%%
\begin{abstract}
Learning from demonstration for motion planning is an ongoing research topic.  
In this paper we present a model that is able to learn the complex mapping from raw 2D-laser range findings and a target position to the required steering commands for the robot.
To our best knowledge, this work presents the first approach that learns a target-oriented end-to-end navigation model for a robotic platform.
The supervised model training is based on expert demonstrations generated in simulation with an existing motion planner.
We demonstrate that the learned navigation model is directly transferable to previously unseen virtual and, more interestingly, real-world environments.
It can safely navigate the robot through obstacle-cluttered environments to reach the provided targets.
We present an extensive qualitative and quantitative evaluation of the neural network-based motion planner, and compare it to a grid-based global approach, both in simulation and in real-world experiments.
\end{abstract}

\section{Introduction}
\label{sec:introduction}

One of the major challenges in robotics is to make robots perform as desired by human operators.
Regarding ground robot navigation, this problem is defined as getting from the current position to a target position, fulfilling the desired navigation policy.
Although objectives like, e.g.~short path or a safe distance to obstacles are perfectly clear to the human operator, it typically requires time-consuming hand tuning such that the robot moves as desired and required.
Additionally, classical motion planning solutions require several steps of data preprocessing that typically are decoupled~\cite{lavalle2006planning}. 
A map of the environment has to be provided, the sensor data has to be preprocessed and potential objects have to be detected such that the planner can react accordingly in a later stage. 

With the aim of reducing the amount of hand-tuning parameters of several processes in order to achieve the desired navigation performance, in this work we present an approach that goes vice-versa: 
\emph{A data-driven end-to-end motion planner}.
The robot is provided with expert demonstrations of how to navigate in a given virtual training environment. 
Like this, a robot operator can show the desired behavior and navigation strategy to the robot.
% does not have to worry on how to make the robot behave in the desired way.
%Given the sensor data, the local target information and the motion commands provided by the expert operator, our approach is able to learn an end-to-end motion planner for local collision avoidance while driving to the target.
During navigation, the goal is not only to replicate the provided expert demonstrations in one specific scenario as in teach-and-repeat approaches \cite{furgale2010visual}, but rather to be able to learn collision avoidance strategies and transfer them to previously unseen real-world environments.

\begin{figure}[tbp]
  \centering
  \frame{\includegraphics[width=\columnwidth]{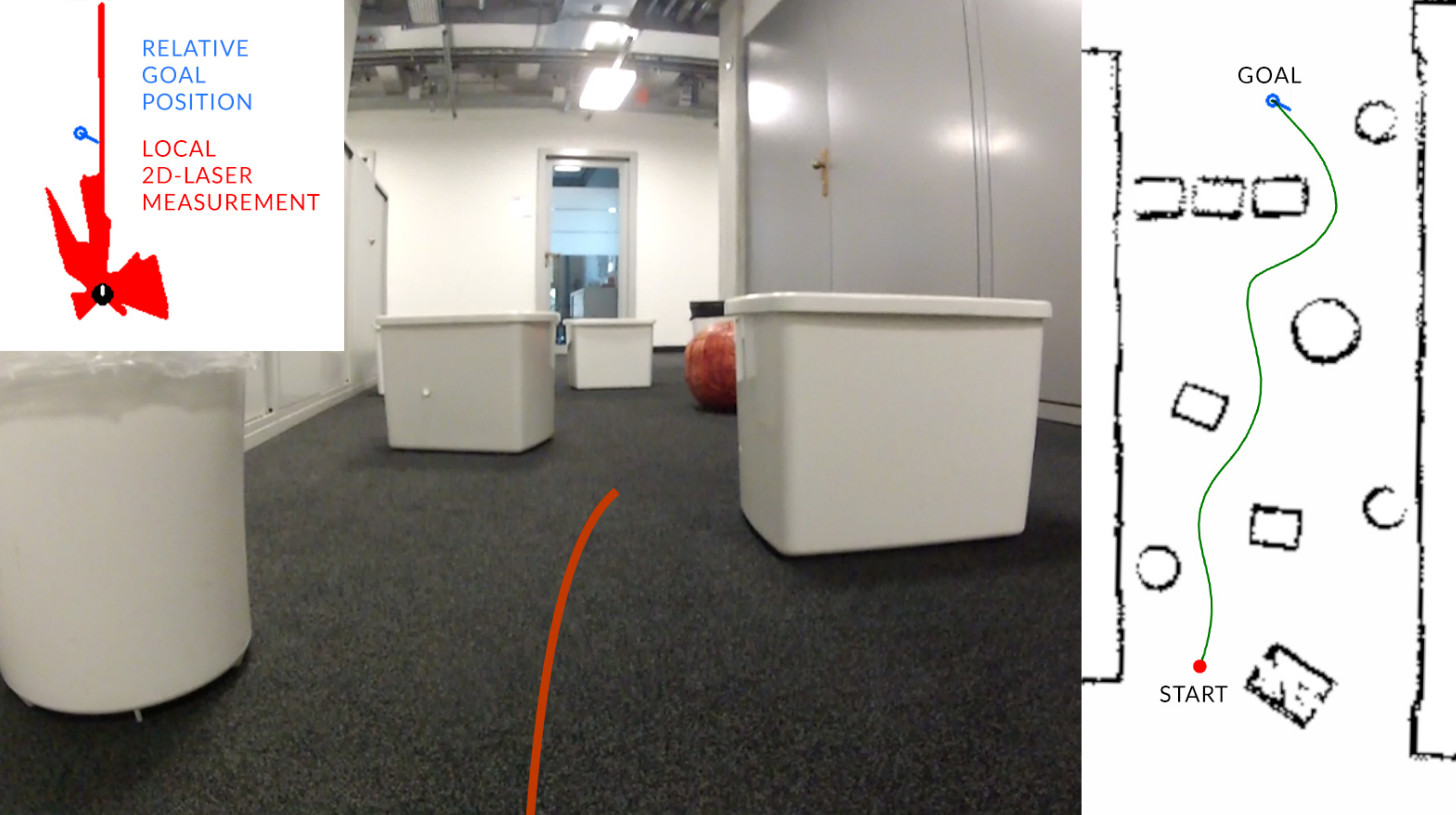}}
\caption{The robotic platform is able to safely navigate through a maze-like environment only using local laser and target information (upper left).
The final traversed trajectory is visualized on the map (right).
The robot view image is only shown for visualization purposes but not used as an input to the end-to-end algorithm.}
\label{fig:teaser}
\vspace{-5mm}
\end{figure}

In comparison to multi-layer map based approaches as presented in \cite{lavalle2006planning}, our approach does not require a global map to navigate. 
Given the sensor data and a relative target position, the robot is able to navigate to the desired target while avoiding collisions with surrounding obstacles.
The approach is constructed under the hypothesis that by learning a physical understanding of the environment and the navigation characteristics of an expert operator, our machine learning-based approach is able to perform in a similar way, even in previously unseen scenarios.
By design, the approach is not limited to any kind of environment. 
However in this work, our analysis is focused on the navigation in static environments.

Computing the motion commands directly from the laser data can be an arbitrarily complex task which requires a model that is able to capture the relevant characteristics for local motion planning. 
Among various machine learning approaches, \ac{dnn}-based ones have the largest potential to model complex dependencies.
They have shown their potential for complex physical scene understanding and feature extraction in various applications like \cite{fragkiadaki2015billiard,punjani2015deep,cadenaMAE-RSS16}, to name a few. % In this work we will focus to model the end-to-end dependency with neural networks.
To train an end-to-end motion planner, we make use of an existing global path planning approach that provides complete trajectories from start to goal positions.
Since for spatial scene understanding and collision avoidance the distances to surrounding objects are of special interest, we use a front facing \SI{270}{\degree} laser range finder as the only sensor of the robot.
%During the supervised learning procedure 
The model is trained on simulation data generated with a global path planner as the expert operator.

The performance of the learned end-to-end motion model is tested both in simulation and on a real robotic platform.
In order to analyze the generalization error of the model for local motion planning, it is especially important to conduct experiments in previously unseen environments.
Therefore, we deployed the robot for evaluations in unknown environments which additionally were significantly more challenging than the one used for training.

The main contributions of this work are: 
\begin{itemize}
  \item A data-driven end-to-end motion planner from laser range findings to motion commands.
  \item Deployment and tests on a real robotic platform in unknown environments.
  \item Extensive evaluation and comparison of the presented local approach with respect to a motion planner with global map information.
\end{itemize}

The remainder of this document is structured as follows: 
In \Cref{sec:related_work} we present an overview of the related work. 
\Cref{sec:approach} formulates the problem and presents our approach.
The experiments and their results are shown in \Cref{sec:experiments} before we discuss the results in \Cref{sec:discussion} and draw a conclusion in \Cref{sec:conclusion}.

\section{Related work}
\label{sec:related_work}

Data driven end-to-end motion planning covers various research areas, both from the perception and the motion planning side.
On the perception side, the scene understanding part is especially important. 
The input data has to be processed to extract relevant information.

%Regarding semantic scene understanding based on image data, various applications showed that it is possible to build an abstract model of the environment shown in images~\cite{socher2011parsing, farabet2013learning}.
% like presented by Socher \textit{et al.}~\cite{socher2011parsing} and Farabet \textit{et al.}~\cite{farabet2013learning} already showed that it is possible to build an abstract model of the environment shown in images.
Since collisions with surrounding objects need to be avoided and the target has to be reached, in our work especially the physical or more precisely the spatial scene understanding has to be given.
Fragkiadaki \textit{et al.}~\cite{fragkiadaki2015billiard} showed that it is possible to learn a model for ball-ball and ball-wall dynamics based on demonstrations shown to the model during training time.
Without any prior knowledge about ball kinematics and collisions, this model is able to predict the motion of several billiard balls in previously unseen configurations.
It shows that it is possible to model physical/spatial interactions using \ac{dnn}s. 
Chen \textit{et al.}~\cite{chen2015deepdriving} presented an approach that extracts spatial information from image data and uses the extracted features for motion planning of an autonomous car.
The approach is not end-to-end since it consists of multiple processing layers, yet it already shows that the extracted information can be used directly by a motion planner.
Ondurska \textit{et al.}~\cite{ondruska2016deep} presented an end-to-end application of neural networks for dynamic object tracking using laser data.
Their work shows that neural networks can be used to extract important spatial information from \ac{2d} laser data.

On the motion planning side, previous work already showed the performance gain by using learning-based motion models for navigation.
Abbeel \textit{et al.}~\cite{Abbeel2008} showed the application of apprenticeship learning for learning human's navigation strategies on a parking lot.
The application was able to significantly reduce the amount of hand-tuning for motion planning.
However, knowledge about the map and road network is required which makes the approach specific for a single environment.
The approaches by Kuderer \textit{et al.}~\cite{kuderer2012rss}, Pfeiffer \textit{et al.}~\cite{pfeiffer2016iros} and Kretzschmar \textit{et al.}~\cite{kretzschmar2016socially} use the maximum entropy inverse reinforcement learning techniques to interaction-aware motion models.
The learned interaction models for pedestrians are based on demonstrated behaviour in occupied environments.
%In such scenarios, the chosen trajectories by pedestrians are highly dependent on interactions with others and therefore it is important to find accurate models that can be used during motion planning.
These applications also show that the amount of hand-tuning of the motion- and interaction models can be reduced by applying machine learning techniques and that the learned motion models can outperform hand-engineered ones.

For applications that require a close coordination between perception and control, like path planning for a robotic arm for closing a bottle, Levine \textit{et al.}~\cite{levine2016end} showed that end-to-end learning approaches can be beneficial and outperform multi-layered ones.
A policy search problem is transformed into a deep reinforcement learning problem that uses a \ac{cnn} for the complex mapping between states and actions.
The learned model can successfully plan motor torques of a robotic arm, given raw image data.
Another image-based deep reinforcement learning approach for motion planning/decision making was presented by Mnih \textit{et al.}~\cite{mnih2015human}.
They showed that it is possible to learn to play various computer games based on screen pixel data and even to outperform human players.

Regarding mobile robotic applications of end-to-end learning, Ross \textit{et al.}~\cite{ross2013learning} presented an approach that learns a left/right controller for an \ac{uav} based on image data.
The \ac{uav} was able to autonomously navigate through a forest while successfully avoiding collisions with trees in the majority of the cases.
However, only the left/right motion has to be controlled while the forward motion command is still selected by a human operator.
Kim \textit{et al.}~\cite{kim2015deep} extended this approach to a hallway application where they learn translational and rotational velocities.
Yet the approach was only tested in empty hallways without any objects blocking the \ac{uav}'s path.
Another image-based end-to-end collision avoidance approach is presented by Muller \textit{et al.}~\cite{muller2005off}.
They focus on the image feature extraction and the transferability among different environmental conditions.
It is shown that the collision avoidance works, yet the navigation performance of the robot is not analyzed.

Sergeant \textit{et al.}~\cite{sergeant2015autoencoders} presented a laser-based and data-driven end-to-end motion planning approach based on deep auto-encoders. 
The collision avoidance capabilities of this approach are shown in simulation and on a robotic platform.
Unlike our framework, no robot target position is taken into account and therefore it is not applicable as a local motion planning technique if a target has to be reached.
With this approach the robot can drive reasonable paths, however no specific goal can be reached.

\section{Approach}
\label{sec:approach}

This section describes the underlying problem and our approach to solve it.

\subsection{Problem formulation}
\label{sec:problem_formulation}

Humans have outstanding capabilities in perceiving the environment, extracting important information and taking rational decisions based on this information.
However, to take decisions, they can rely on a large knowledge base gained throughout their entire lives.

In order to take the right decisions --- in our application this is to move a robot to the desired target position (including heading) --- robots have to overcome several challenges.
First, the relevant information has to be extracted from the sensor data.
Second, using this information, a model has to be found that describes the relationship between the observations and the actions to take.
Third, during deployment, this model has to be used to take the right decisions as soon as new observations are available.

Whereas many approaches solve these tasks independently, we present an approach that directly computes suitable steering commands based on sensor and target data.
Given expert demonstrations, we try to find a function
\begin{equation}
\u = \F_{\params} \big (\y, \g \big)
\end{equation}
that directly maps a vector of sensor data $\y$ and goal information $\g$ to desired steering commands $\u$. 
This function is parametrized by a parameter vector $\params$.
During supervised training we find the function parameters $\params$ that best explain a set of training data.
The optimization criterion is based on $| \F_{\params} (\y, \g) - \u_{\mathrm{exp}}|$, the difference between the predicted steering commands and the ones provided by the expert operator.
During deployment, the model parameters $\params$ are given and the steering commands can be computed given the input data $\y$ and $\g$.

\subsection{End-to-end model}
\label{sec:model}
The end-to-end relationship between input data and steering commands can result in an arbitrarily complex model.
Among different machine learning approaches, \ac{dnn}s/\ac{cnn}s are well known for their capabilities as a hyper-parametric function approximation to model complex and highly non-linear dependencies.
In this work, we will use a \ac{cnn} for the representation of the function $\F_{\params}$.
The entire processing pipeline of extracting information and finding the right steering commands has to be covered by a single model.

\begin{figure}[t]
  \centering
  \includegraphics[width=\columnwidth, trim=0 0 0 0, clip]{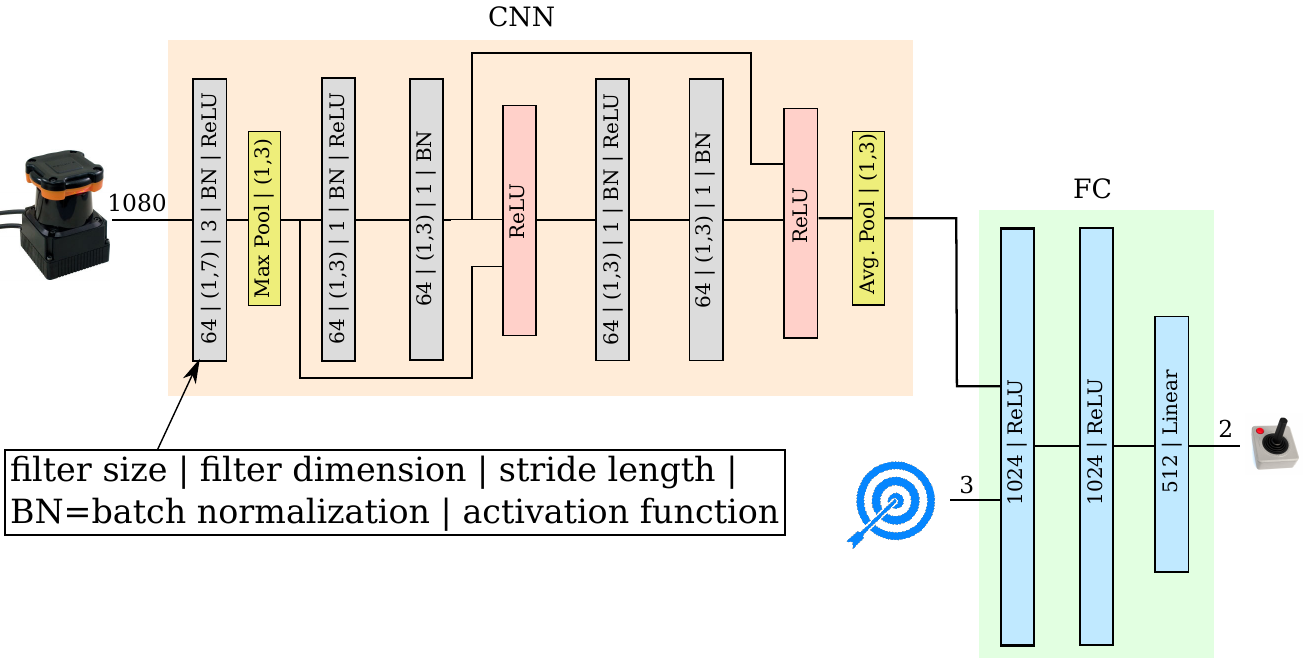}
  \caption{Structure of the \ac{cnn}.
  The laser data is processed by the convolutional part which consists of two residual building blocks as presented in \cite{he2015deep}.
  The \ac{fc} part of the network fuses the extracted features and the target information.
  The input/output dimensions of the overall model are shown on the connections.
  $L_1$ regularization is applied to all model parameters.}
  \label{fig:nn_structure}
  \vspace{-5mm}
\end{figure}

As mentioned above, the inputs are given by the measurements of the \ac{2d} laser range finder and the relative target position which means the position of the target (polar coordinates) in a robot-centric coordinate system.
In order to retrieve spatial scene understanding features, the laser data is processed by a \ac{cnn} before the outputs of that sub-network are fused with the target information and processed by the \ac{fc} layers of the model.
The structure of the neural network model is shown in \Cref{fig:nn_structure}.
The network consists of two residual building blocks including shortcut connections as suggested in \cite{he2015deep}, where it was shown that the training complexity can be reduced by using residual networks, compared to stacked convolutions.

Throughout this paper, two versions of the model will be investigated. 
For the first version (\textit{CNN\_smallFC}), the three \ac{fc} layer dimensions are (256, 256, 256) while for the second version (\textit{CNN\_bigFC}), their dimensions are increased to (1024, 1024, 512).
The convolutional part of both networks remains unchanged, as shown in \Cref{fig:nn_structure}.
Our neural network model implementation is based on Google's TensorFlow framework \cite{tensorflow2015-whitepaper}.

\subsection{Model training}
The ultimate goal for the presented approach is to be able to learn a driving characteristic demonstrated to a robot by an expert operator in a supervised manner. %, e.g. a human or another motion planner.
To avoid the burden of human driving data collection at big scale, we resort to simulation where a global motion planner is used as an expert.
This is a very valuable feature in robotic applications where data collection is expensive.
For each time step $i$, the data tuple $\gamma_i = (\y_i, \g_i, \u_{\mathrm{exp}, i})$ consists of laser measurements, target information and an expert velocity command $\u_{\mathrm{exp}, i}$. 
Here, the velocity command $\u = (v, \omega)$ includes translational and rotational velocity.
In order to reduce temporal correlations in the training data to a minimum, the tuples are randomized before used for training. 
The optimization is conducted using the \emph{Adam} optimizer \cite{adam2015iclr} with mini-batch training.
The loss function for each supervised learning step $k$ is given by

\begin{equation}
J_{k} \big ( \Gamma_B \big) = \frac{1}{N_B} \cdot \sum_{j=i}^{i+N_B} | \F_{\params_{k}} (\y_j, \g_j) - \u_{\mathrm{exp}, j}| , 
\end{equation}

\noindent where the mini-batch $\Gamma_B = [\gamma_i, \dots, \gamma_{i+N_B}]$ is comprised of multiple samples of the training data tuples. 
$\F_{\params_{k}}$ represents the model at the current training step.
The gradient of this cost function w.r.t. the model parameters can be computed using backpropagation.

\subsection{Motion planner deployment}
One advantage of neural network models compared to other approaches is their predictable query time during testing. 
Whereas the complexity of multi-stage approaches might increase if the environment becomes more complicated, the complexity of a \ac{dnn} is unaffected by the robot's environment and the query remains unaltered. 
Since no external preprocessing of the laser data is required, the computational complexity and therefore also the query time for a steering command only depends on the complexity of the model, which is constant once it is trained.

The presented neural network model computes the steering commands frame-by-frame.
No internal or external memory is used to take into account previous in- and outputs.

\section{Experiments}
\label{sec:experiments}

This section covers the conducted experiments and their evaluation. 
First, the robotic platform is introduced.
It is used throughout all experiments, both as a model in simulation and for the real-world tests.
Second, the training data generation is explained before four experiments (two in simulation, two on the real platform) for evaluation are presented.
In the following, the neural network based motion planner will also be referred to as the deep planner.

\subsection{Robotic platform}
We use a Kobuki based TurtleBot as a robotic platform.
We added a front-facing Hokuyo UTM\footnote{https://www.hokuyo-aut.jp/02sensor/07scanner/utm\_30ln.html} laser range finder (see video\footnote{https://youtu.be/ZedKmXzwdgI}) with a \ac{fov} of \SI{270}{\degree} and a maximum scanning range of \SI{30}{\meter} to the differential drive robot.
The angular resolution of the laser sensor is \SI{0.25}{\degree} which leads to 1080 measurements.
We use an Intel$^{\circledR}$ NUC with an i7-5557U processor with \SI{3.10}{\GHz} running Ubuntu 14.04 as an onboard computer and the \ac{ros} \cite{ros} as a middleware.
 
%\begin{figure}[ht]
%  \centering
%  \includegraphics[width=0.4\columnwidth, trim=0 0 0 0, clip]{turtlebot}
%  \caption{Kobuki TurtleBot equipped with a front-facing \ac{2d} Hokuyo laser range finder and an Intel$^{\circledR}$ NUC unit.}
%  \label{fig:turtlebot}
%  \vspace{-5mm}
%\end{figure}

\subsection{Training data}
\label{sec:training-data}
As in \cite{sergeant2015autoencoders}, we use the \ac{ros} \ac{2d} navigation stack for the expert two-level motion planner (global and local).
On the global level we use a grid-based Dijkstra \cite{lavalle2006planning} planner while a \ac{dwa} \cite{fox1997dynamic} planner is used on the local level.
% The global costmap uses the robot radius $r_R = $ \SI{0.178}{\meter} and an inflation radius $r_{inf.} = $ \SI{0.5}{\meter} with a cost scaling factor of 10.
% The local planner is a \ac{ros} internal dynamic window planner\footnote{\url{http://wiki.ros.org/dwa\_local\_planner}} \cite{fox1997dynamic} with the parameters provided in \Cref{tab:planner-params}.
% \setlength{\belowcaptionskip}{-5pt}
% \begin{table}[H]
% \caption{Parameters used for the local planner with the cost function weights for path\_distance\_bias (PDB), goal\_distance\_bias (GDB) and occdist\_scale (ODS).}
% \centering
% \label{tab:planner-params}
% \setlength\tabcolsep{1.5 pt}
% \begin{tabular}{|c|c|c|c|c|c|c|c|}
% \hline
%   $v_{min}$ & $v_{max}$ & $|\omega_{max}|$ & $|\dot v_{max}|$ & $|\dot \omega_{max}|$ & PDB & GDB & ODS\\ \hline \hline
%  \SI{0}{\meter/\second} & \SI{0.6}{\meter/\second} & \SI{\pi}{\radian/\second} & \SI{2 }{\meter/\second^2} & \SI{2}{\radian/\second^2} & 70 & 24 & 1 \\ \hline
% \end{tabular}
% \end{table}
% \setlength{\belowcaptionskip}{15pt}
Stage \ac{2d} is used as a dynamic simulator.

During training data generation, the robot drives to randomly selected target positions on the $\SI{10}{\meter} \times \SI{10}{\meter}$ training map (\textit{train}).
The simulated sensor data, relative target positions and expert steering commands are recorded.
The selected target positions are guaranteed to be collision free, e.g. lie outside of possible obstacles.
The original \training map is shown in the particular leftmost column of \Cref{fig:error-boxplot} and \Cref{fig:trajectory_comparison}, respectively.
For our real-world tests, we used fused training data both from the \training and the \evalcomplex map.
The latter comprises clutter and other objects the robot might face during real-world tests.
We generated 6000 trajectories in the \training map and 4000 trajectories in the \evalcomplex map with 2.1M and 2.2M input/output tuples, respectively.
Training the model on a Nvidia GeForce GTX 980 Ti GPU\footnote{http://www.geforce.com/hardware/geforce-gtx-980-ti/buy-gpu} roughly takes \SI{8}{\hour} with the conducted 2M training steps.

There were four reasons that caused the decision to only use simulation data (generated with an existing planner) for training:
\begin{enumerate*}[label=(\roman*)]
\item The data generation is deterministic as the \ac{ros} planner is deterministic,
\item for basic experiments we can eliminate noisy sensor data as a potential reason for failure,
\item it is faster and easier to generate data and
\item we can test the robustness and the generalization capabilities of the approach if we apply a model trained in simulation on a real robot.
\end{enumerate*}

\subsection{Frame-by-frame evaluation}
\label{sec:frame-evaluation}
The experiment focuses on the evaluation error of the \ac{cnn} model regarding the computed steering commands.
As mentioned in \Cref{sec:training-data}, the model was trained with the samples from the \training map only. 
This experiment is conducted using the \textit{CNN\_smallFC} model.
In order to be independent of a GPU during testing, the query has to be done using a CPU only.
On an Intel$^{\circledR}$ i7-4810MQ with \SI{2.8}{\GHz} the average model query time is \SI{4.3}{\milli \second}.

For the evaluation, we generated input/output tuples for the three maps --- comprising \textit{train}, \evalsimple and \evalcomplex --- by driving to 30 random target positions each, using the expert motion planner.
Given this data (unseen during training), the error between the translational and rotational steering commands of the deep and the expert motion planner is computed for each input/output tuple.

\begin{figure}[t]
  \centering
  % Plot was generated with the script compare_maps_for_model.py
   \begin{overpic}[scale=1, trim={0 0.0cm 0 0.2cm}, clip]{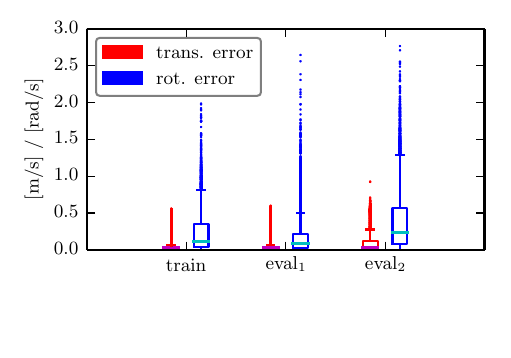}
     \put(29,0){\includegraphics[scale=0.038]{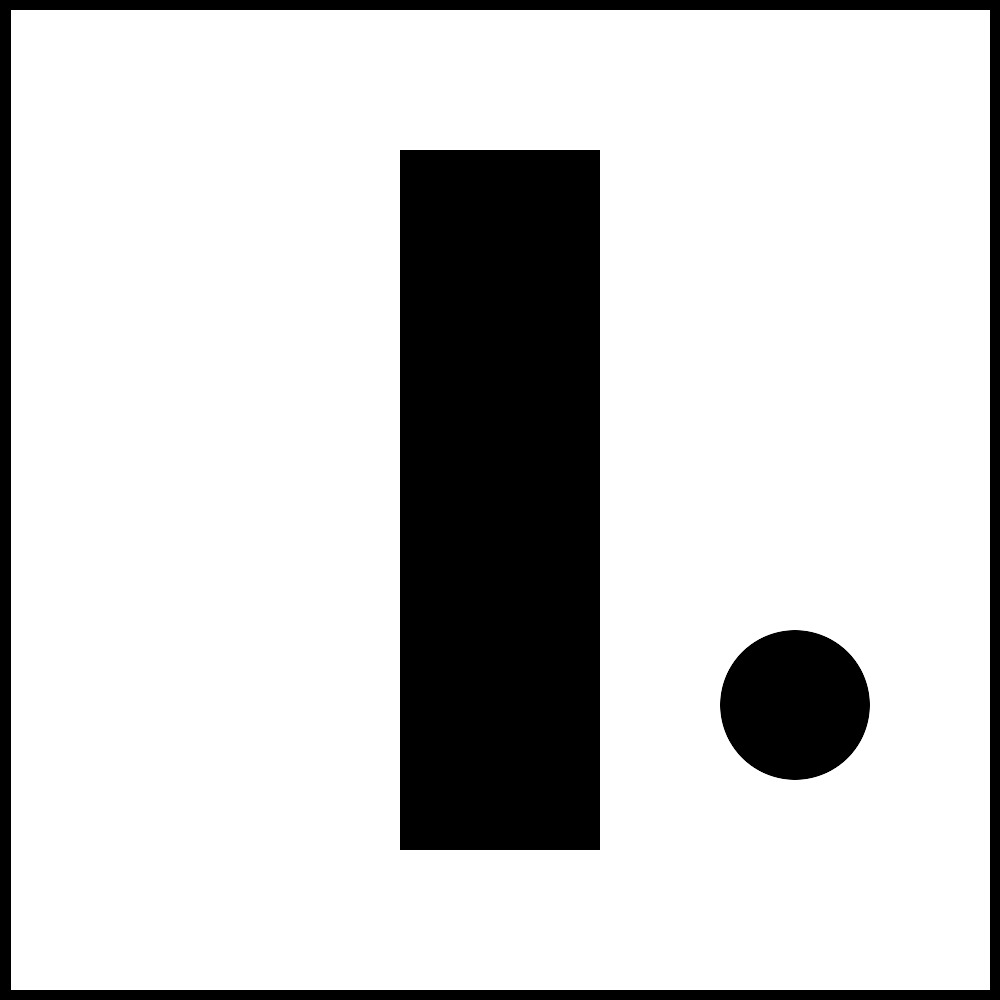}}
     \put(49,0){\includegraphics[scale=0.038]{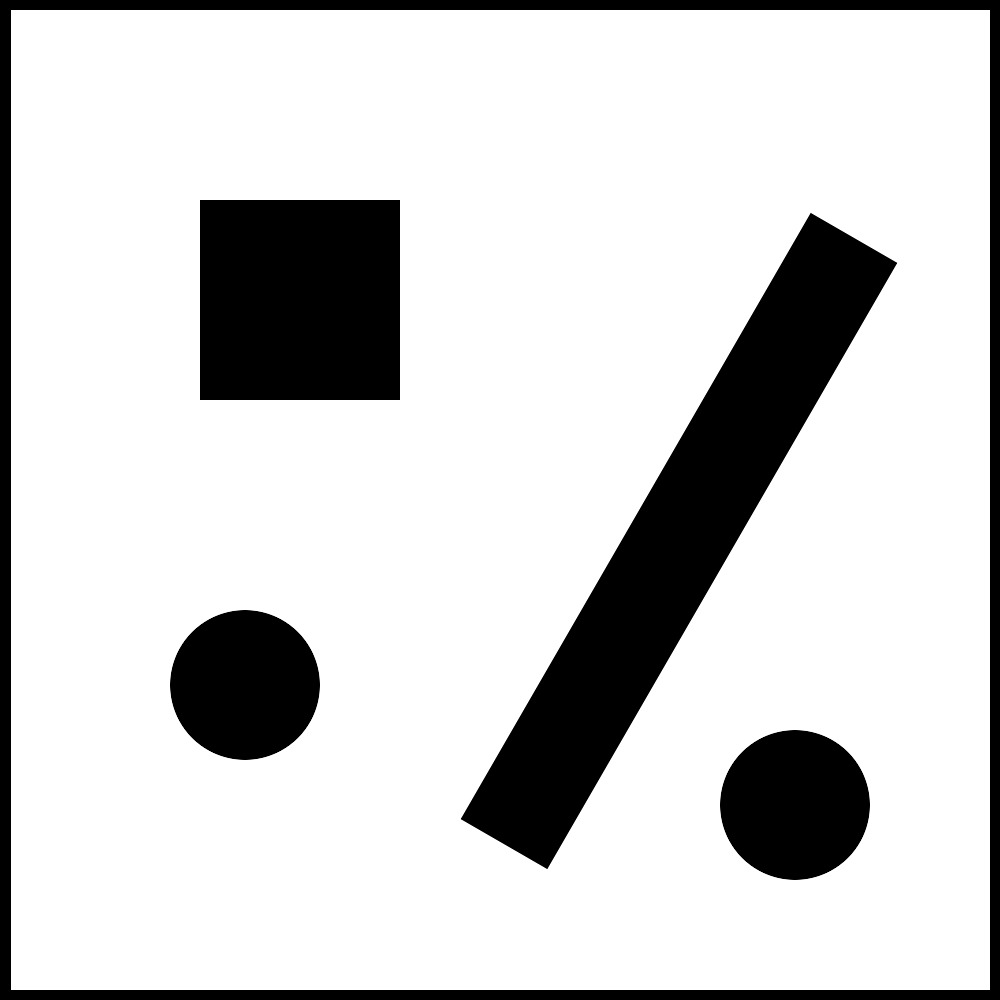}}
     \put(69,0){\includegraphics[scale=0.038]{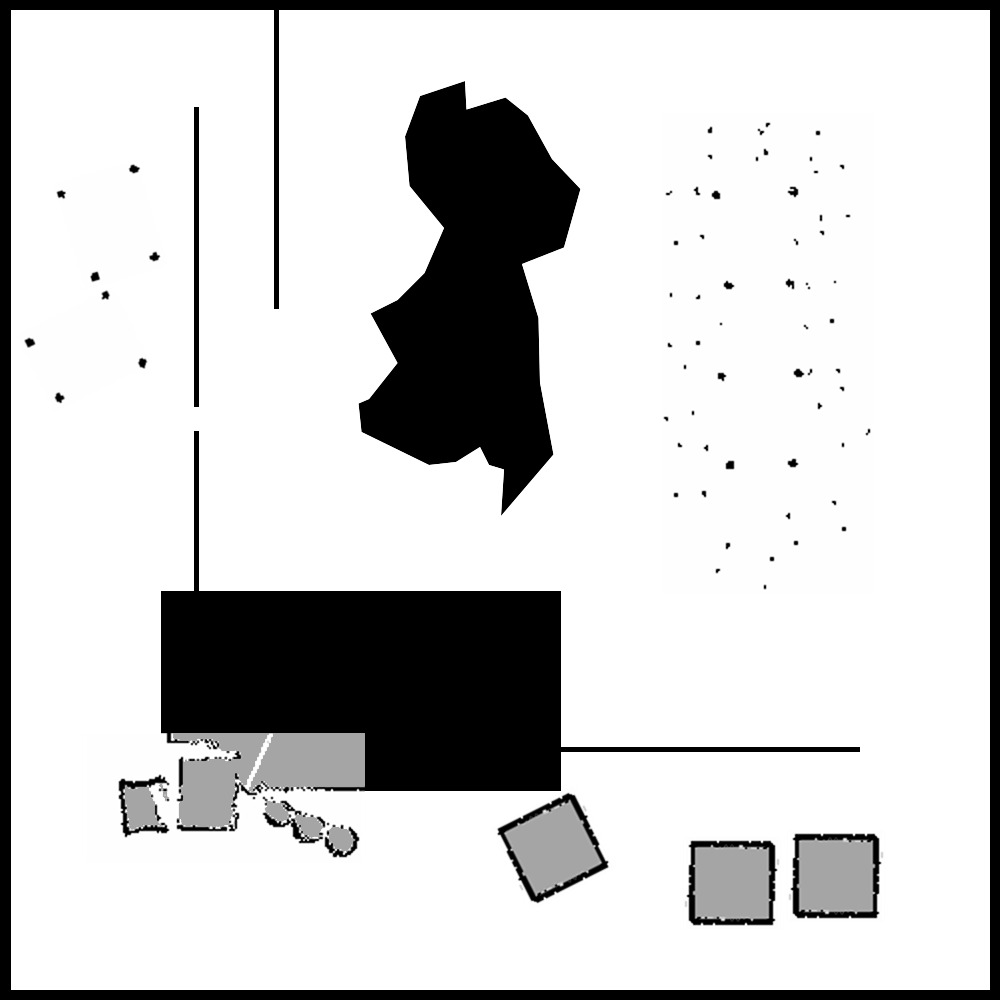}}
  \end{overpic}
\caption{Error statistics of the frame-by-frame error between the \ac{ros} (expert) and the deep planner.
 The evaluation data was not used for training before. 
 The three small figures visualize the maps on which the evaluation was conducted.
 Maps are better viewed by zooming in on a computer screen.}
\label{fig:error-boxplot}
\vspace{-5mm}
\end{figure}
As the error statistics in \Cref{fig:error-boxplot} suggest, the smallest evaluation error can be observed on the \training map. 
We identified that the large outliers of the rotational velocity command typically occur when turning on the spot. 
For example turning \SI{180}{\degree} either to the right or left has a large impact on the rotational velocity error but only a minor impact on the actual robot behavior.
The \evalsimple map has a different structure compared to the \training map, yet the obstacles have a similar shape.
This causes an increase of the evaluation error, especially the rotational velocity command.
Moving on to the \evalcomplex map, both the translational and the rotational part of the evaluation error increase further.

This result confirms our expectations. 
The model is able to transfer the scene understanding and navigation knowledge gained in one environment to another one. 
However, if the deviation in the structure of the environment is large --- as e.g. between the \training and the \evalcomplex environment --- the proper scene understanding might not be given which leads to the larger deviation between expert- and deep planner steering commands.

\subsection{Trajectory comparison in a simulated environment}
\label{sec:traj-comp-sim}
The previous experiment showed that it is possible to transfer learned knowledge from one map to another, yet the results were only analyzed frame-by-frame.
In this section we want to analyze the performance of the navigation model when it is deployed as a motion planner on our robotic platform.
This experiment shows, whether the navigation characteristics were learned from the expert or whether there was overfitting on a specific map.
The \ac{cnn} model and training data are the same as in the previous experiment.

\begin{figure}[t]
  \centering
  % Plot was generated with the python script plot_missions.py
  \includegraphics[width=\columnwidth, trim={0 0.0cm 0 0}, clip]{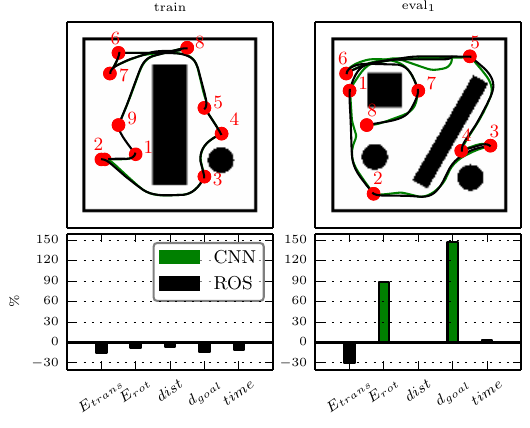}
  \vspace{-7mm}
  \caption{Performance comparison between the \ac{ros} (expert) and the deep planner.
  Testing results of the trained navigation model on the \training (left) and \evalsimple map (right).
  Top: comparison on a trajectory level with target locations marked in red.
  Bottom: relative error in [\%] of the \ac{ros}/deep planner for the evaluation metrics: final distance to goal ($d_{goal}$), translational energy ($E_{trans}$), travelled distance ($dist$), rotational energy ($E_{rot}$) and the travel time ($time$).
  Green bars indicate a relative error of the deep planner with respect to the \ac{ros} expert, black bars the opposite.
  Therefore, positive (green) error means that the \ac{ros} planner is better, negative (black) error means that the deep planner is better.}
  \label{fig:trajectory_comparison}
  \vspace{-5mm}
\end{figure}

Both for the \training and the \evalsimple environment, missions with fixed target positions are created.
%For each environment, both planners have to drive to the targets provided by the given mission.
While the \ac{ros} planner has global knowledge about the map as during training, the deep planner only receives the relative target (red dots, \Cref{fig:trajectory_comparison}) and the laser range findings at each timestep as an input.
Since the simulation is deterministic, only one mission per planner is evaluated.
The reproducability of the approach will be tested in the succeeding real-world experiment.

Since the planning structure used for training is a layered motion planning approach, it cannot be described in a single cost function.
Therefore, in addition to the visual inspection, the trajectories are also evaluated based on the following metrics inspired by \cite{xu2012real}:
\begin{itemize}
  \item $d_{goal}$: distance of the final trajecory position to the given goal point summed over all trajectories
  \item $E_{trans}$: integrated absolute value of translational acceleration over all trajectories
  \item $E_{rot}$: integrated absolute value of rotational acceleration over all trajectories
  \item $dist$: overall travelled distance for the mission
  \item $time$: overall travel time for the mission
\end{itemize}

\Cref{fig:trajectory_comparison}(top) shows an example of the final executed trajectories.
The relative errors between both planners for each of the selected metrics are shown in \Cref{fig:trajectory_comparison}(bottom).
Those performance metrics are independent of the cost function of the map-based \ac{ros} planner but should provide an estimate of the trajectory characteristics of both planners.

The results show that our end-to-end approach is able to capture the navigation policy of the expert operator.
Especially on the \training map, the trajectories driven with both planners are congruent for a majority of the cases, although only local information is used for the deep planner.
This is somehow expected, given that the training data was recorded on this map. 
The deep planner slightly outperforms the expert planner in terms of the given metrics.
Since neither the \ac{ros} nor the deep planner were trained or tuned for those metrics, this has nothing to do with overfitting. 
It just rates the performance based on the independent metrics.

Transferring the deep planner to the \evalsimple environment shows how well the knowledge gained from one map is transferable to the other. 
Although at many positions there are different topologies on the map, e.g. whether an obstacle should be passed on the left or right side, the deep planner takes similar actions and routes as the expert planner that has perfect knowledge of the full map.
Regarding the positioning of the robot at the goal positions and the rotational energy, the deep planner now is inferior to the expert one with respect to the given evaluation metrics.
The increase in the rotational energy was already indicated by the rotational velocity error increase in \Cref{fig:error-boxplot}.
Furthermore, this is indicated by the trajectories of the deep planner on the \evalsimple map in \Cref{fig:trajectory_comparison}(top-right), where it shows deviations from the expert trajectories at several positions.
In any of the deviations, the deep planner swerves back to the path of the expert planner, yet the correction causes the extra amount of rotational energy.
Inspite the small deviations the travel distance and time are still similar to the expert planner since as \Cref{fig:trajectory_comparison}(top-right) shows, the deep planner has a slight tendency to cut edges sharper than the \ac{ros} planner.

The figures in the right column of \Cref{fig:trajectory_comparison} clearly show that the performance of the deep planner is worse than on the \training map.
Yet they also confirm that the \ac{cnn} model is able to learn a given navigation characteristic of an expert operator and transfer it to a previously unseen environment and not only to replicate expert demonstrations.

\subsection{Real-world navigation}
\label{sec:real-world-driving}
\begin{figure}[htbp]
  \centering
   \begin{overpic}[width=\columnwidth, trim={0 0cm 0 0cm}, clip]{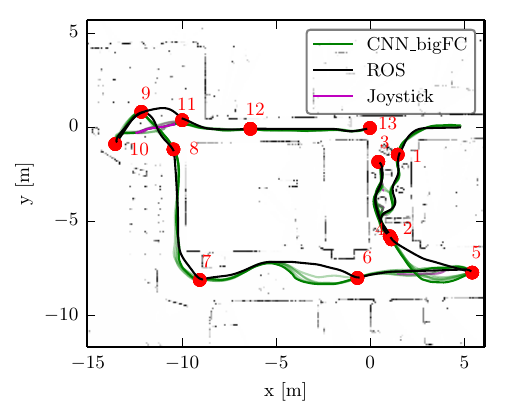}
     \put(79,38){\includegraphics[scale=0.034]{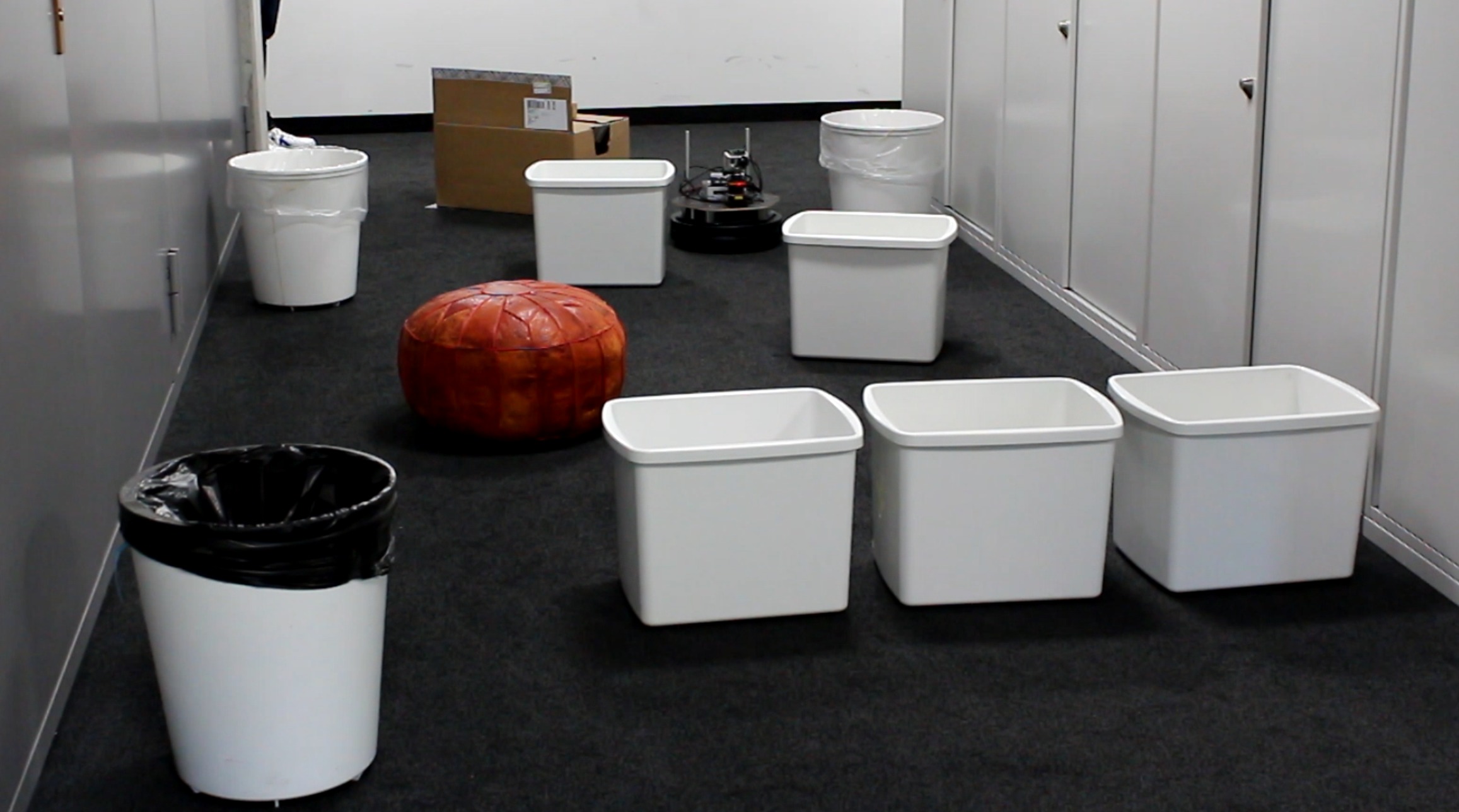}}
     \put(53,13){\includegraphics[scale=0.05]{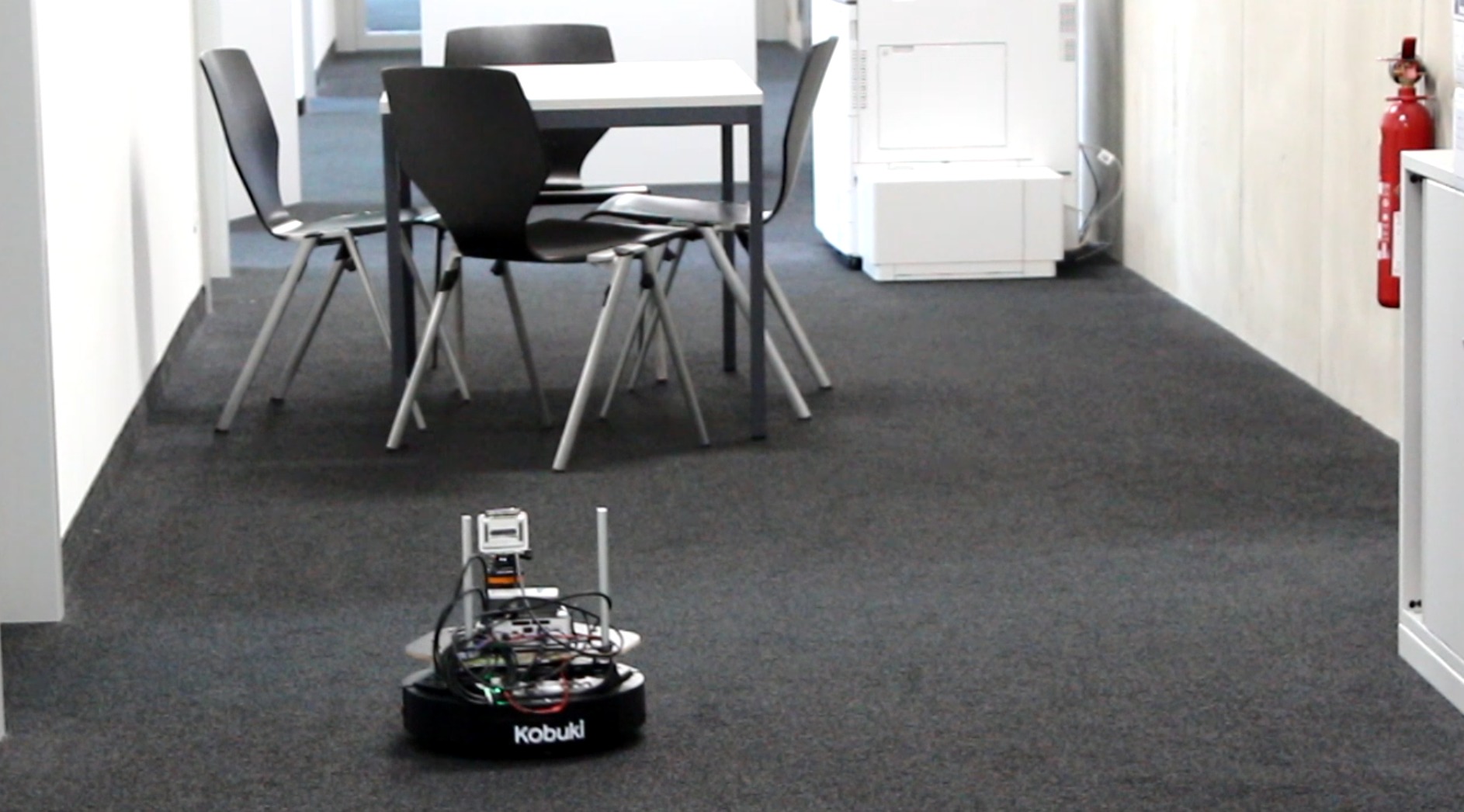}}
     \put(20,64){\includegraphics[scale=0.035]{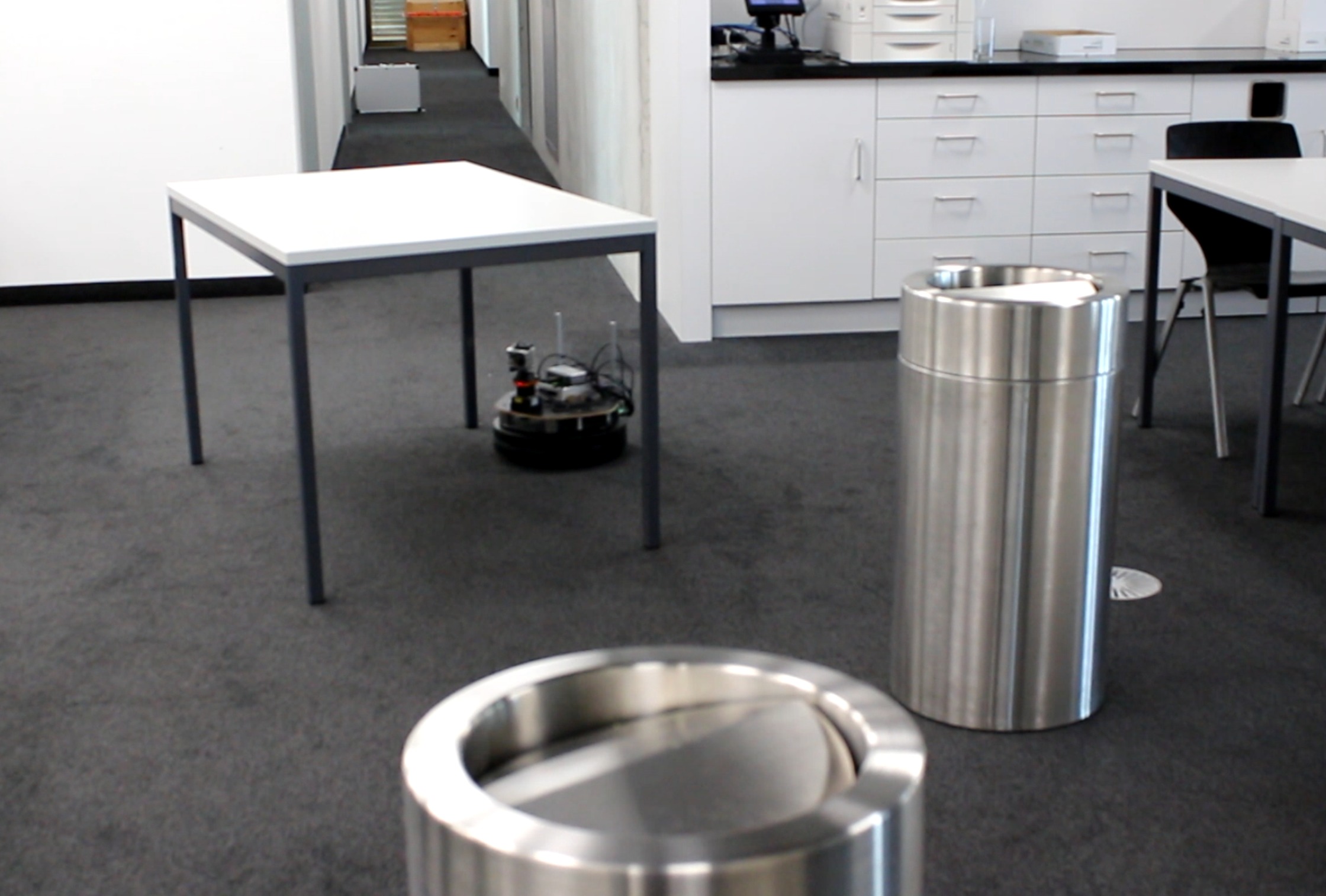}}
  \end{overpic}
\caption{Comparison between the driven trajectories of the \ac{ros} (expert) and the deep planner (\textit{CNN\_bigFC}) on a real robot.
For the expert planner, one experiment is shown while for the deep planner 6 experiments are shown.
The three small pictures give an impression of the actual environment.}
\label{fig:real-driving}
\vspace{-5mm}
\end{figure}

%For the previous experiments, the navigation model was trained and tested in simulation.
The following experiment is similar to the one presented in \Cref{sec:traj-comp-sim}, yet now the driving tests are conducted using the real robot.
In order to be able to localize and to navigate with the expert planner, a map of the environment was recorded beforehand.
As up to now, the navigation with the deep planner is only based on local laser and target information during test time.
During the experiment, the robot has to traverse a maze like area with many obstacles, a table with chairs blocking one half of a corridor, an area with a lot of clutter and also long corridors. 
Snapshots of the environment are shown together with the results of this experiment in \Cref{fig:real-driving} and in the video attachment.
\setlength{\belowcaptionskip}{5pt}
\begin{table}[htbp]
\caption{Amount of manual joystick control needed for successful navigation (in [\%] of the total distance traveled.)}
\centering
\label{tab:completed-targets}
\setlength\tabcolsep{3 pt}
\begin{tabular}{|l|c|c|c|c|c|c|c|c|c|c|c|c|c|}
\hline
 Target 			& 1 & 2 & 3 & 4 & 5 & 6 & 7 & 8 & 9 & 10 & 11 & 12 & 13\\ \hline \hline
 \textit{CNN\_smallFC} & \cellcolor{yellow} 1 & 0 & \cellcolor{yellow}1 & 0 & \cellcolor{red}4 & 0 & 0 & 0 & \cellcolor{red}2 & \cellcolor{red}4 & \cellcolor{red}6 & 0 & 0 \\ \hline
 \textit{CNN\_bigFC} & 0 & 0 & \cellcolor{yellow}1 & \cellcolor{yellow}1 & 0 & \cellcolor{yellow}1 & 0 & 0 & 0 & 0 & \cellcolor{red}4 & \cellcolor{yellow}1 & 0 \\ \hline
\end{tabular}
\end{table}
\setlength{\belowcaptionskip}{15pt}

As mentioned above, for the real-world experiments we trained the model based on the \training and the \evalcomplex environment in order to provide it with the basic navigation principles but also with object shapes that could potentially be observed in reality.
We also found that increasing the size of the \ac{fc} layers of the network improves the navigation performance in the real-world while it was not beneficial during the simulation tests.

The deep planner is able to drive the majority of the missions fully autonomous. 
However, in some positions, the human operator briefly had to ``help'' the robot when it got stuck.
The joystick interventions are marked in \Cref{fig:real-driving}.
\Cref{tab:completed-targets} shows that the amount of joytick interventions was reduced significantly by increasing the size of the three \ac{fc} layers.
Since the environment is richer and more complex, also the extracted features from the \ac{cnn} layers might be more diverse.
The \ac{fc} layers of the \textit{CNN\_smallFC} model most likely are to small to deal with the increased amount of information.

In order to test the repeatability of the deep planner results, we conducted six drives with the \textit{CNN\_bigFC}. 
All of those trajectories are shown in \Cref{fig:real-driving}.
As in simulation, the navigation characteristics of the deep planner are similar to the one of the expert planner.
Here, the deep planner reacts differently in a few areas, however still consistent between the missions.
For example between targets 4 and 5, the turn with the deep planner is wider and between 6 and 7 the deep planner reacts later to the tables than the \ac{ros} planner does.
This is due to the fact that only local information is used while the expert planner has global knowledge of the environment.
Since the laser beams point radially from the robot, the distance to the legs of the table and chairs needs to be sufficiently small such that they are understood as obstacles by the \ac{cnn} model.
Between targets 10 and 11 the required human interference was the highest for both models.
Considering that the robot was only trained in closed environments with relatively well-arranged obstacles, it was not able to traverse this cluttered area fully autonomously in any of the experiments.

Although the joystick interventions were required to reach some target positions, we have to mention that the majority of the interventions were required to ``unstuck'' the robot and not to avoid an imminent collision.
In general, throughout the whole experiment, no instable motion of the robot could be observed.
Interestingly, situations the robot could not handle rather caused it to stop and stand still instead of driving unpredictable and instable paths.

\subsection{Reaction to sudden changes}
\label{sec:non-static}
This last experiment qualitatively shows the performance of the deep planner when facing suddenly appearing and disappearing obstacles.
The results for this experiment are shown in \Cref{fig:dynamic-obj} and in the attached video.
Although the navigation model only computes a single steering command, the constant velocity path for this steering command is shown for visualization purposes.

The robot has to drive from one end of a corridor to the opposite end.
While the path to the target is clear in the beginning (\Cref{subfigure:straight}), the robot is faced with a suddenly appearing object blocking its path (\Cref{subfigure:avoid}).
The deep planner clearly reacts to the object by swerving to the right. 
After removing the obstacle, the robot corrects its path in order to approach the target as fast as possible (\Cref{subfigure:straight_again}).

\setlength{\belowcaptionskip}{-5pt}
\setlength{\subImageWidth}{0.3\columnwidth}
\begin{figure}[htbp]
  \centering
  \begin{minipage}{\subImageWidth}
	\includegraphics[width=\subImageWidth, trim={0 0 0 0cm}, clip]{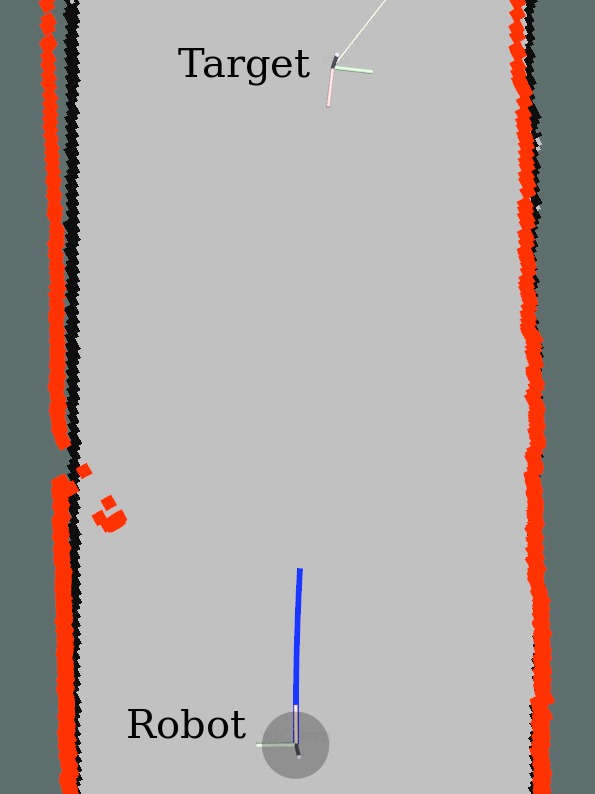}
	\vfill
	\includegraphics[width=\subImageWidth, trim={0 0cm 0 0cm}, clip]{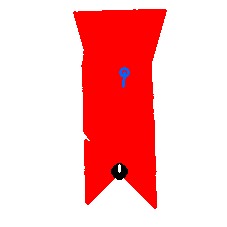}
	\vfill
	\includegraphics[width=\subImageWidth, trim={0 3cm 0 3cm}, clip]{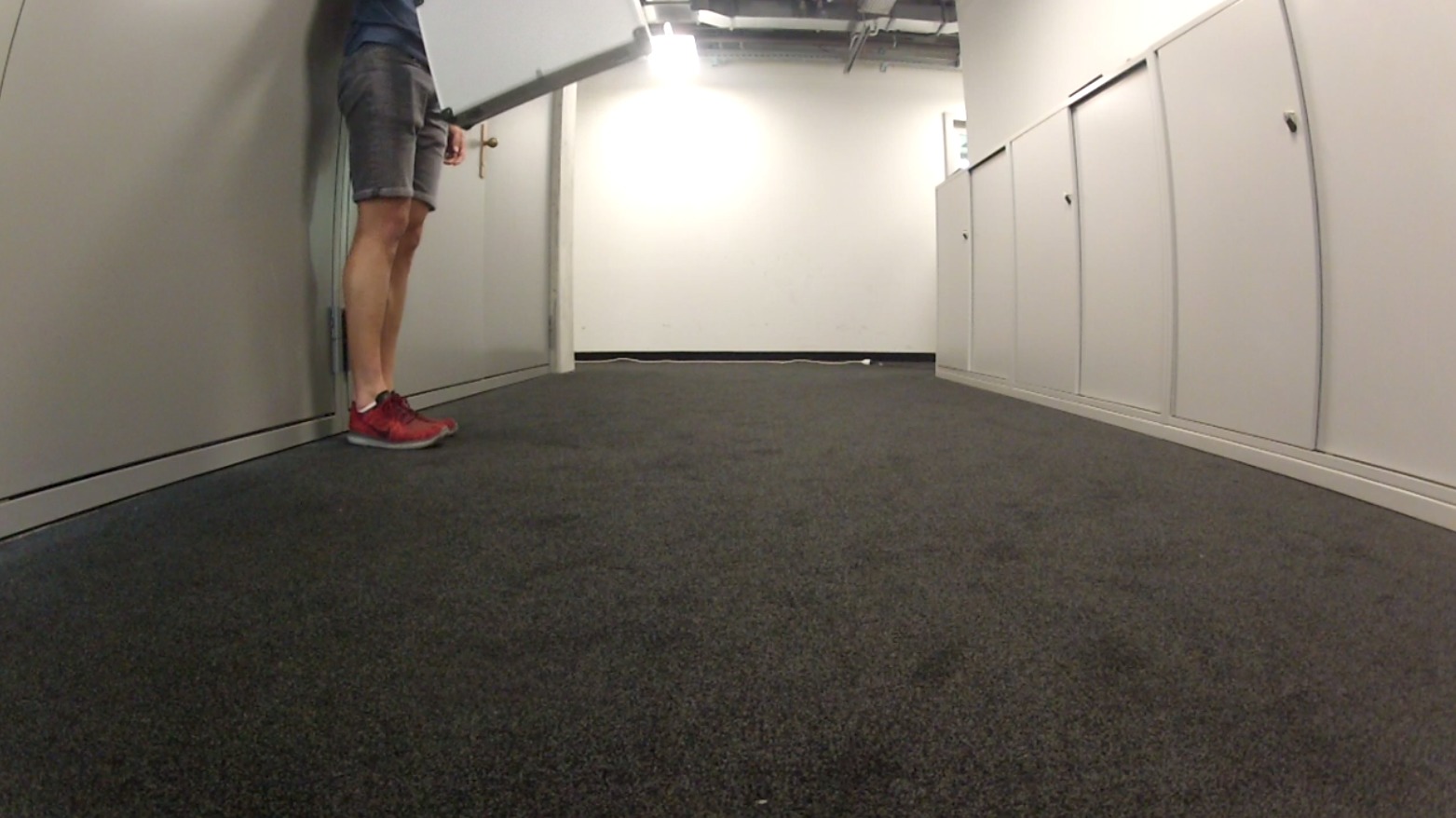}
	\subcaption{}
	\label{subfigure:straight}
  \end{minipage}
  \begin{minipage}{\subImageWidth}
	\includegraphics[width=\subImageWidth, trim={0 0 0 0cm}, clip]{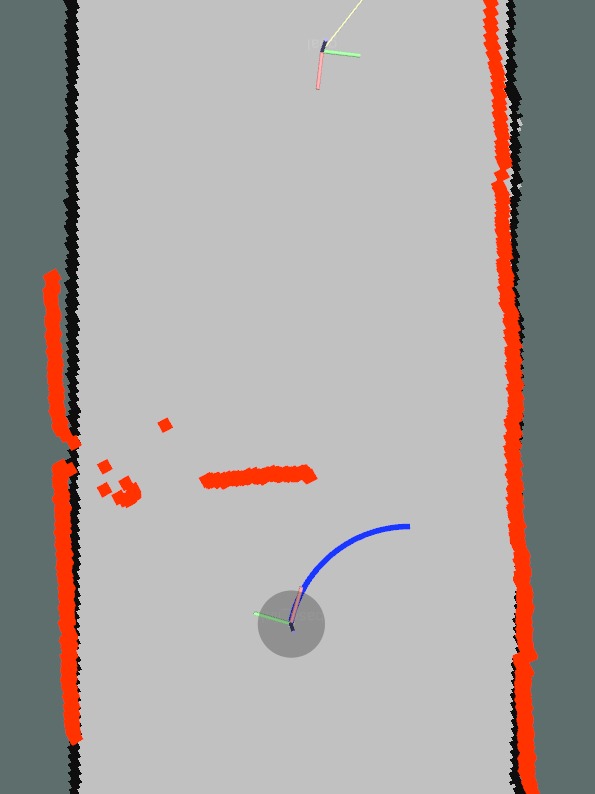}
	\vfill
	\includegraphics[width=\subImageWidth, trim={0 0cm 0 0cm}, clip]{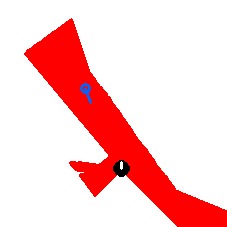}
	\vfill
	\includegraphics[width=\subImageWidth, trim={0 3cm 0 3cm}, clip]{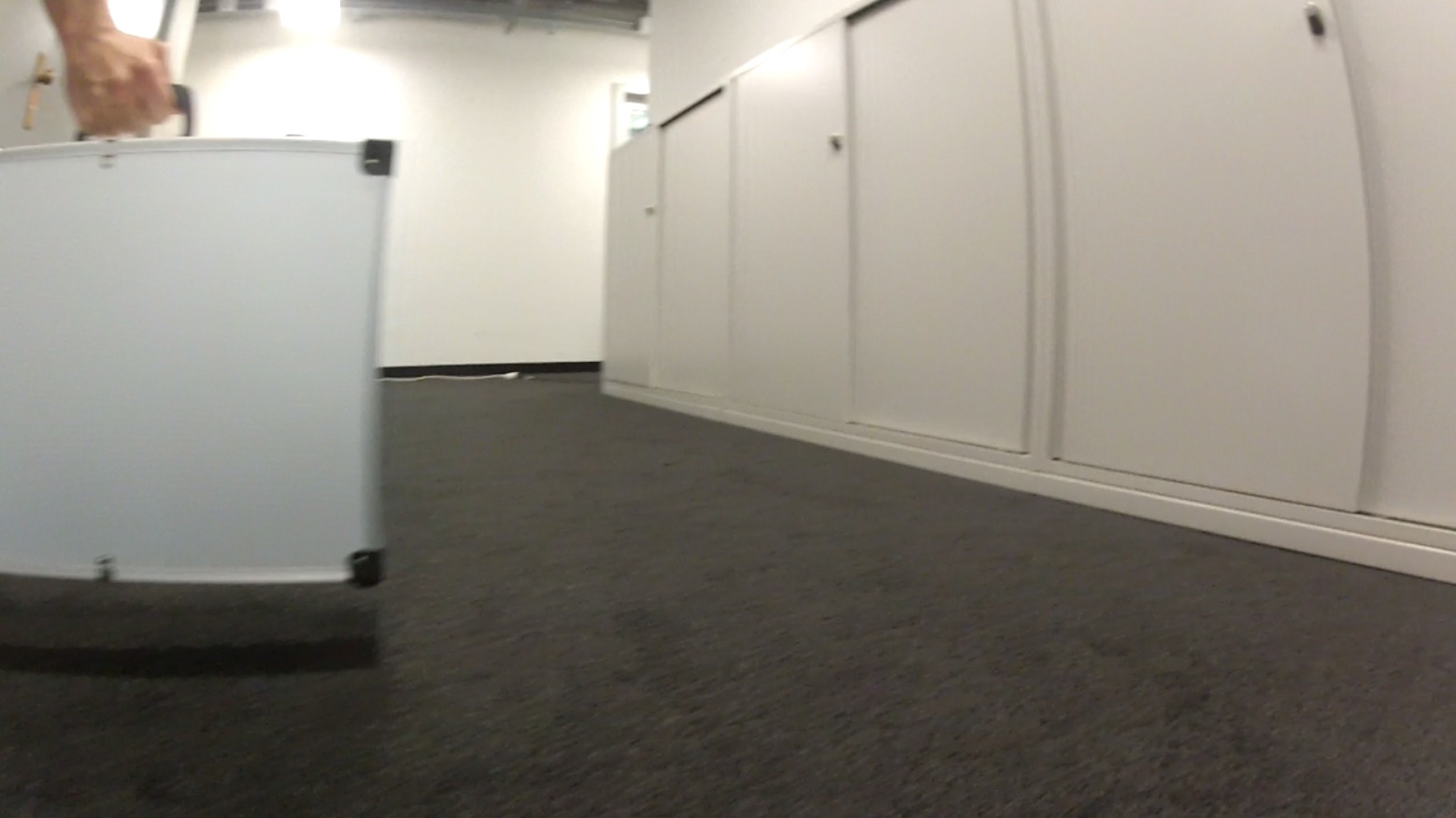}
	\subcaption{}
	\label{subfigure:avoid}
  \end{minipage}
  \begin{minipage}{\subImageWidth}
	\includegraphics[width=\subImageWidth, trim={0 0 0 0cm}, clip]{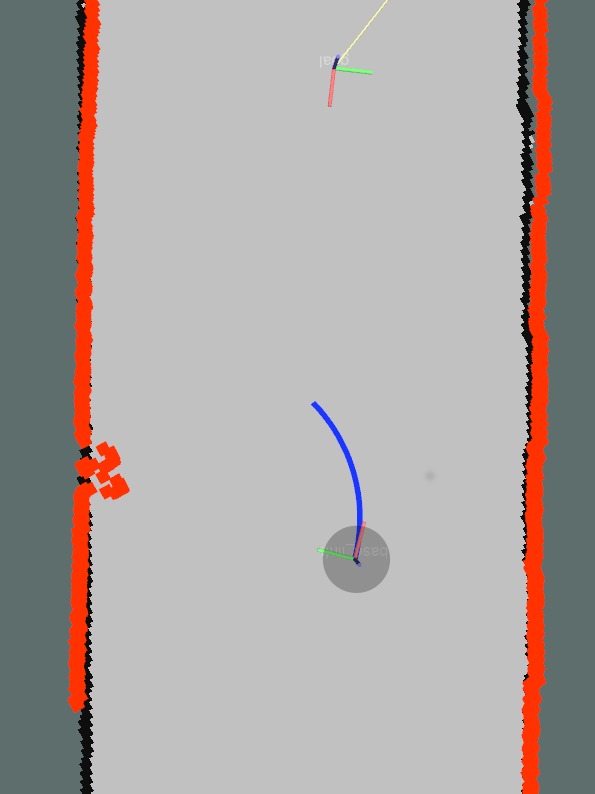}
	\vfill
	\includegraphics[width=\subImageWidth, trim={0 0cm 0 0cm}, clip]{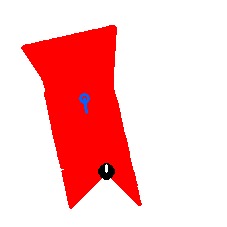}
	\vfill
	\includegraphics[width=\subImageWidth, trim={0 3cm 0 3cm}, clip]{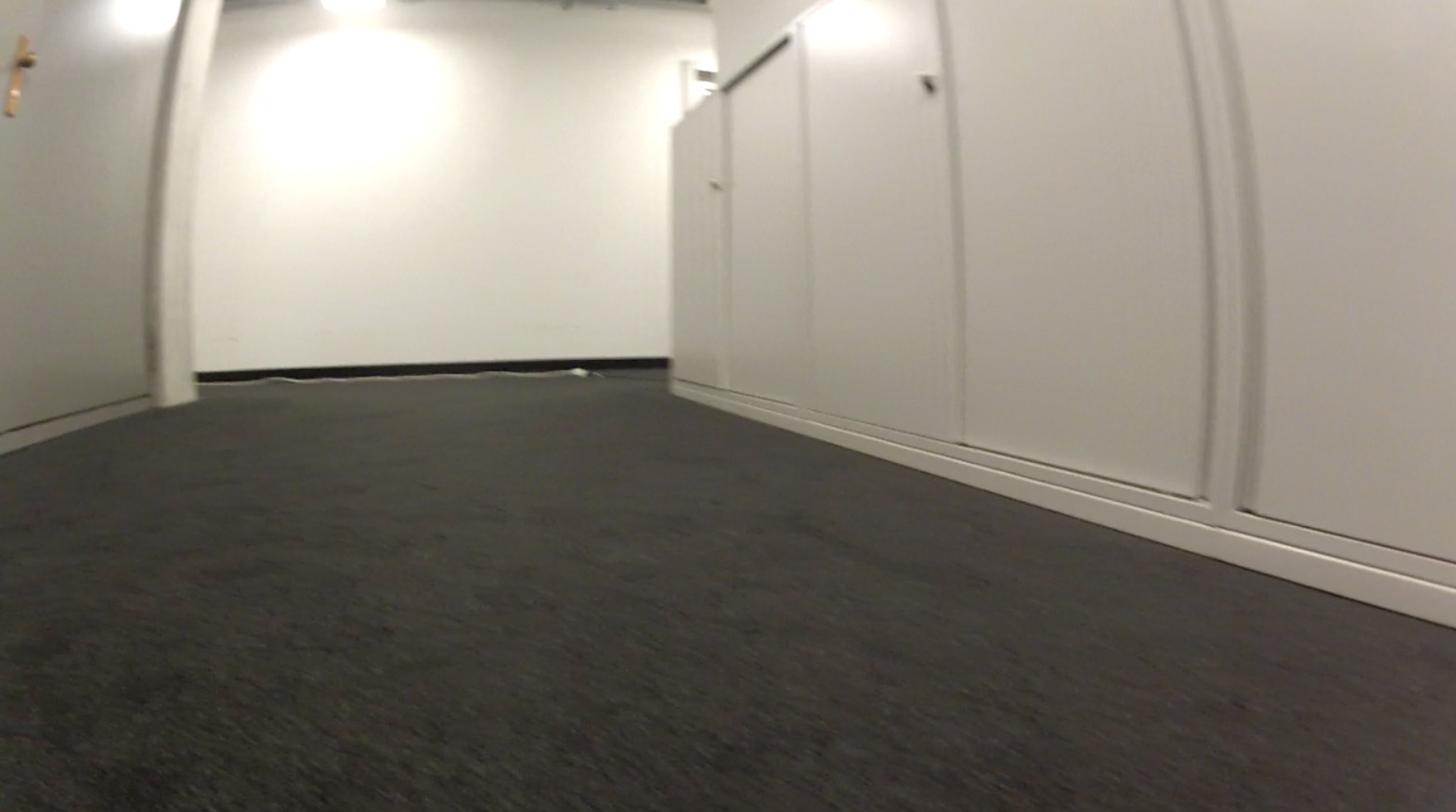}
 	\subcaption{}
	\label{subfigure:straight_again}
  \end{minipage}
	\setlength{\belowcaptionskip}{15pt}
  \caption{Reaction to unforeseen objects. 
  While driving towards the target position, the path of the robot gets blocked by an obstacle and afterwards it is freed again.
  In the top row of this figure, the constant velocity path using the computed steering commands and the overall setup are visualized. 
  The middle row visualizes the robot-centric laser range findings and the relative target position fed to the robot while robot-view images of the last row are only used for visualization purposes.}
	\label{fig:dynamic-obj}
\vspace{-5mm}
\end{figure}

\section{discussion}
\label{sec:discussion}
In this work we showed that it is possible to learn a navigation policy from an expert operator making use of a \ac{cnn} architecture for the complex end-to-end mapping function from raw sensor data to steering commands.
We showed in various experiments that the learned navigation model is able to transfer gained knowledge from training environments to unseen and complex environments.
One important finding of this work is that it is possible to train such a model in simulation and deploy it on a real platform while still having satisfactory navigation performance.
This in as extremely valuable feature in many robotic applications where data generation is expensive.

To our best knowledge, this is the first approach that is able to perform target oriented navigation and collision avoidance based on an end-to-end approach using neural networks.
Our experiments showed that the model is capable of much more than solving trivial start to goal scenarios or simple collision avoidance tasks. 
It is even able to solve complex navigation tasks in maze-like environments in the majority of the cases, only using local information.

Although we compare our approach to a map-based motion planner, we are fully aware that it cannot completely replace a map-based path planner.
If the environment becomes more complex, the deep planner is still limited to be a local motion planner that relies on a global path planner to provide targets.
At this point it also has to be clarified that our goal is not to replicate an existing planner with a \ac{cnn}. 
Ideally, the final training data would come from a human operator that can train or re-train the robot himself.
Another option might be to use training data from an optimal global planner which however is not real-time feasible.
Taking several thousand trajectories from a single human demonstrator might be infeasible. 
Therefore, further research will show how the training samples can be reduced and how simulated and real-world training data can play together.

During the real world experiments, we found that one limitation of the current approach is wide open spaces with a lot of glass and/or clutter around. 
This potentially results from the fact that the model was trained purely from perfect simulation data. 
Training or re-training a model using real sensor data might reduce this effect. 

Furthermore, we observed that the deep planner is able to avoid small dead ends if it approaches them from the outside. 
Once the robot enters a convex dead-end region, it is not capable of freeing itself.
In addition to that, the robot's heading sometimes fluctuates before avoiding an obstacle. 
This issue will be further analyzed and might be solved by using recurrent neural networks with internal memory.

\section{Conclusion}
\label{sec:conclusion}
In this work we presented a data-driven end-to-end motion planning approach for a robotic platform.
Given local laser range findings and a relative target position, our approach is able to compute the required steering commands for a differential drive platform.

The end-to-end model is based on a \ac{cnn} and is able to learn navigation strategies from an expert operator and transfer this knowledge between different environments.
We provided an extensive evaluation for simulation and real-world experiments and showed that it is possible to train a navigation model using simulation data and deploy it on a real robotic platform in an unseen environment.

%%%%%%%%%%%%%%%%%%%%%%%%%%%%%%%%%%%%%%%%%%%%%%%%%%%%%%%%%%%%%%%%%%%%%%%%%%%%%%%%

% Bibliography
\footnotesize
\bibliographystyle{style/IEEEtran}
\balance
\bibliography{bib/IEEEabrv,bib/IEEEfull,bib/deep_planning}

\end{document}